\RequirePackage{fix-cm}
\documentclass[twocolumn]{svjour3}          
\smartqed  
\usepackage{graphicx}
\usepackage[english]{babel}
\usepackage[utf8]{inputenc}

\usepackage[table, dvipsnames]{xcolor}
\usepackage{yaacro}
\usepackage{booktabs}
\usepackage{diagbox}
\usepackage{cite}

\usepackage{soulutf8}

\usepackage{amsmath,amssymb,amsfonts,amsthm}

\usepackage[tight,footnotesize]{subfigure}

\usepackage[colorinlistoftodos]{todonotes}

\usepackage{multirow}
\usepackage{ulem}
\newcommand{\etal}{\textit{et al. }}
\newcommand{\robotloc}{\vec{x_k}}

\newcommand{\state}{\vec{s}}

\DeclareMathOperator{\rmap}{q}

\newcommand{\revision}[1]{\textcolor{black}{#1}}

\begin{document}

\title{Scalable Multirobot Planning for Informed Spatial Sampling}


\author{Sandeep Manjanna$^{1}$ \and M. Ani Hsieh$^{1}$ \and Greogory Dudek$^{2}$
}


\institute{
            \email{msandeep@seas.upenn.edu, mya@seas.upenn.edu, dudek@cim.mcgill.ca}\\
            $^{1}$GRASP Laboratory, University of Pennsylvania, Philadelphia, USA. \\$^{2}$Center for Intelligent Machines, McGill University, Montreal, Canada
}


\date{Received: date / Accepted: date}

\maketitle



\begin{abstract}

This paper presents a distributed scalable multi-robot planning algorithm for informed sampling of quasistatic spatial fields. We address the problem of efficient data collection using multiple autonomous vehicles and consider the effects of communication between multiple robots, acting independently, on the overall sampling performance of the team. We focus on the distributed sampling problem where the robots operate independent of their teammates, but have the ability to communicate their current state to other neighbors within a fixed communication range. 
Our proposed approach is scalable and adaptive to various environmental scenarios, changing robot team configurations, and runs in real-time, which are important features for many real-world applications. We compare the performance of our proposed algorithm to baseline strategies through simulated experiments that utilize models derived from both synthetic and field deployment data. The results show that our sampling algorithm is efficient even when robots in the team are operating with a limited communication range, thus demonstrating the scalability of our method in sampling large-scale environments.

\end{abstract}


\section{Introduction}\label{sec:intro}

In this paper, we present a distributed planning approach to design paths for multiple robots to achieve efficient sampling of a quasistatic spatial field. Our objective is to plan a non-myopic path for the robots to maximize the information gain in a limited time. We address the problem of efficient sampling of environmental processes using multiple autonomous vehicles.

\begin{figure}[t]
	\begin{center}
		\centering
		\includegraphics[width=0.95\columnwidth]{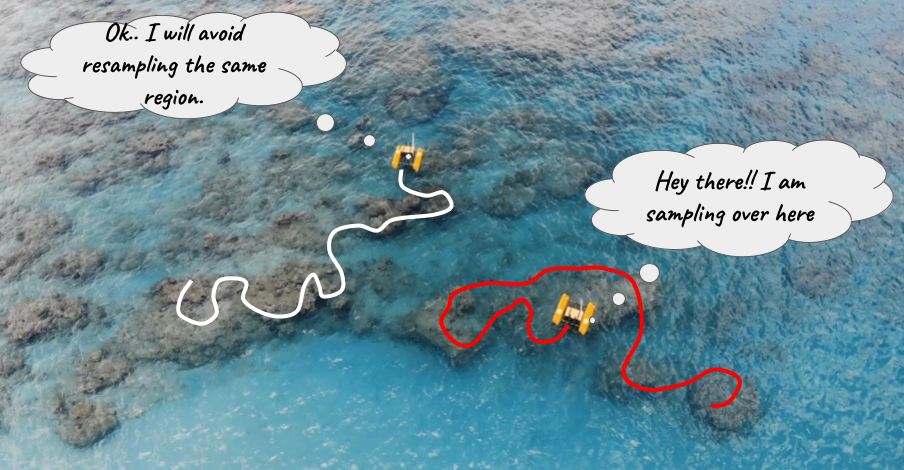}
		\caption{Aerial view of two robotic boats surveying the reef at a marine reserve off the coast of Barbados island. Example paths of the boats are illustrated with solid lines of colors red and white. The goal is to plan paths that are rewarding in terms of information gain and have minimal overlap. We achieve higher rewards by determining the best policy and minimal overlap through communication between the robots.}
		\vspace{-1.8em}
		\label{fig:multi_asv}
	\end{center}
\end{figure}

Many natural processes can be modeled by hotspots exhibiting extreme measurements and higher spatial variability than the rest of the field. Examples of such spatial fields include algal blooms, coral reefs (see Fig.\ref{fig:multi_asv}), distribution of flora and fauna, and aerosol concentration. To model such environmental phenomena with high precision, we need the sampling robots to visit many information rich locations within their limited endurance. One possible way to achieve this is by exploring the hotspot regions in the early stages of the survey as the hotspots are rich with the information needed to model the phenomenon of interest. Estimating a good representation of spatial phenomena plays a key role in applications like environmental monitoring, search and rescue, anomaly detection, and geological surveys. We propose an informed non-myopic path planning technique for robotic platforms to efficiently collect measurements from a quasistatic spatial field, so that an accurate model of the underlying physical phenomenon can be built.

\begin{figure}[ht]
	\begin{center}
		\centering
		\includegraphics[width=0.99\columnwidth]{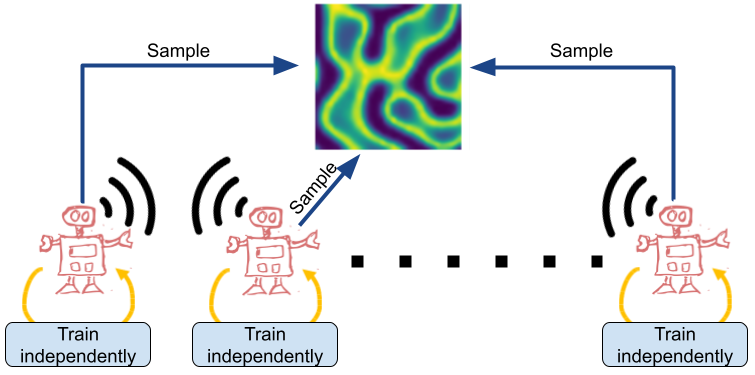}
		\caption{Overview of our approach. We train each individual robot to perform efficient sampling and deploy them together to analyze the coverage behavior when the robots are given an ability to communicate.}
		\vspace{-1.8em}
		\label{fig:overview}
	\end{center}
\end{figure}

In a multi-robot setup, task division between the robots becomes essential as opposed to a single robot sampling approach~\cite{manjanna2018policy}. We do not formulate the multi-robot scenario as a distributed or collective learning problem. Instead, we train each robot as an independent agent\footnote{\revision{The robots are assumed to be homogeneous in terms of their capabilities. They may all have the same set of learnt parameters. However, this is not compulsory as long as they are all trained on similar distributions.}} and observe their sampling performance when they are put together in a sampling task and are given the ability to intermittently communicate as shown in Fig.~\ref{fig:overview}. Doing so makes the problem more tractable; and the multi-robot system becomes easily scalable as the same trained parameters can be used on any new robot added to the system. The team becomes more resilient to failure as all robots are individually trained to perform the task efficiently, and the team can be put together in any robot combination.

\begin{figure}[ht]
	\begin{center}
		\centering
		\includegraphics[width=0.99\columnwidth]{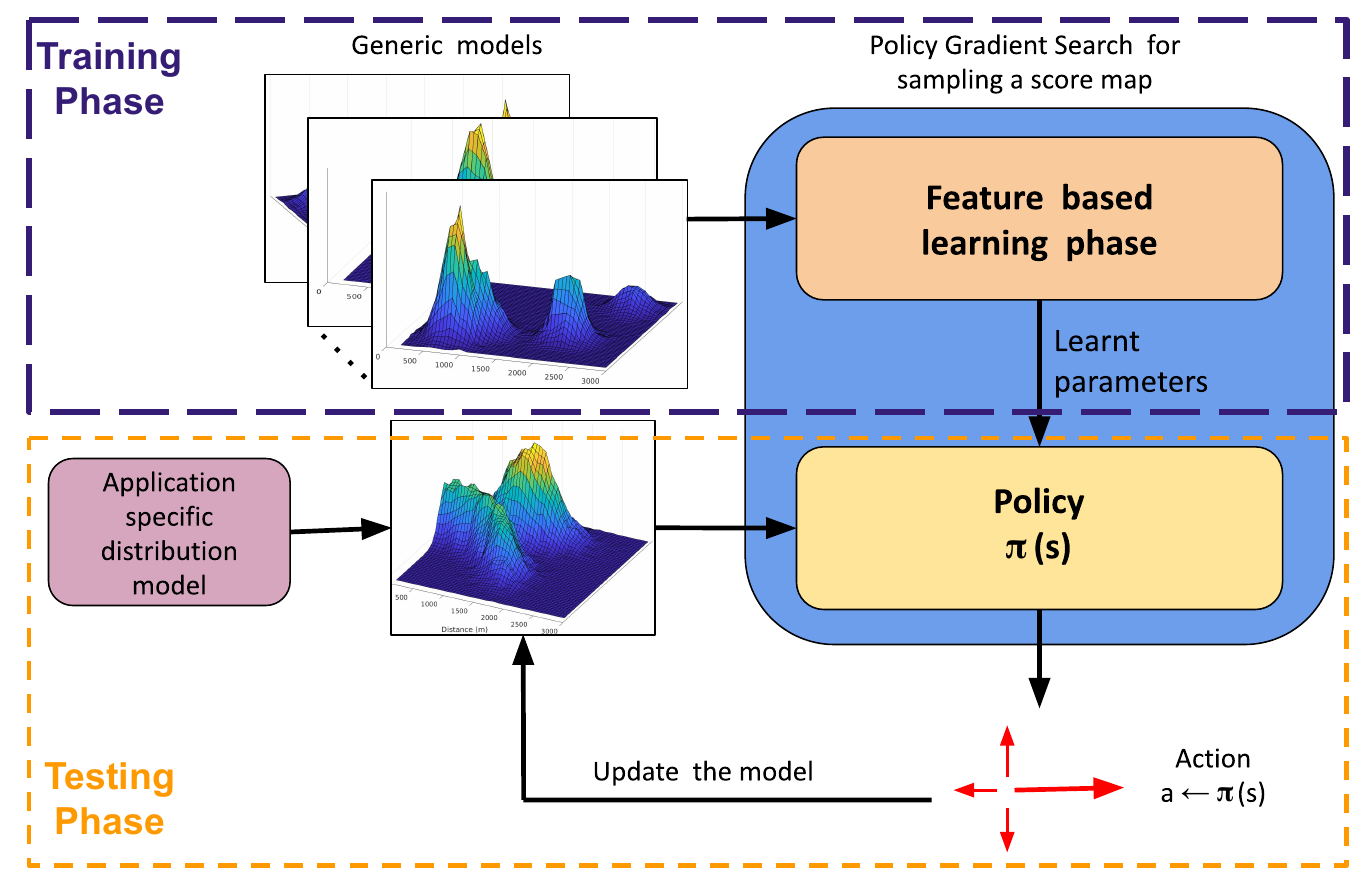}
		\caption{Overview of the training and testing phases.}
		\vspace{-1.8em}
		\label{fig:training_overview}
	\end{center}
\end{figure}

\revision{An overview of the training and testing phases of an individual robot is illustrated with a block diagram in Fig.~\ref{fig:training_overview}. In the training phase, the parameters for the policy are learnt on a set of generic distributions that represent the field that needs to be surveyed. Then in the testing phase, when the robot is being deployed in the field, the prior of the current field is used to generate a policy for the robot given its current location or state. In our example application of sampling visual data of the coral reef using a robotic boat, we train the algorithm with a set of satellite images of different coral regions. While deploying the robot with these learnt parameters from the training, we use an aerial image of the survey area to provide a prior scoremap for the algorithm to generate an action plan for the robotic boat.}

\revision{Once the robots are trained, they are all put together in a sampling task and are allowed to communicate only their visited locations with other neighboring robots that are within the communication range.} Communication between agents plays an important role in a decentralized multi-robot system. In this paper, we consider the effects of communication between multiple robots on the overall sampling performance of the team. We compare the overall sampling performance of the team as the communication range of the robots changes. Our proposed multi-robot sampling algorithm performs efficiently even when robots in the team are operating with a limited communication range, thus demonstrating the scalability of our approach for sampling large-scale environments.

Key contributions of this paper include: 
\begin{itemize}
    \item A scalable multi-robot system with distributed decision making to achieve nonuniform sampling of a quasistatic field.
    \item Analysis of the effect of robot communication range on the overall sampling performance of the team.
    \item Investigation of the task-sharing behavior between robots.
\end{itemize}

\begin{figure*}
	\begin{center}
		\centering
		\includegraphics[width=0.99\textwidth]{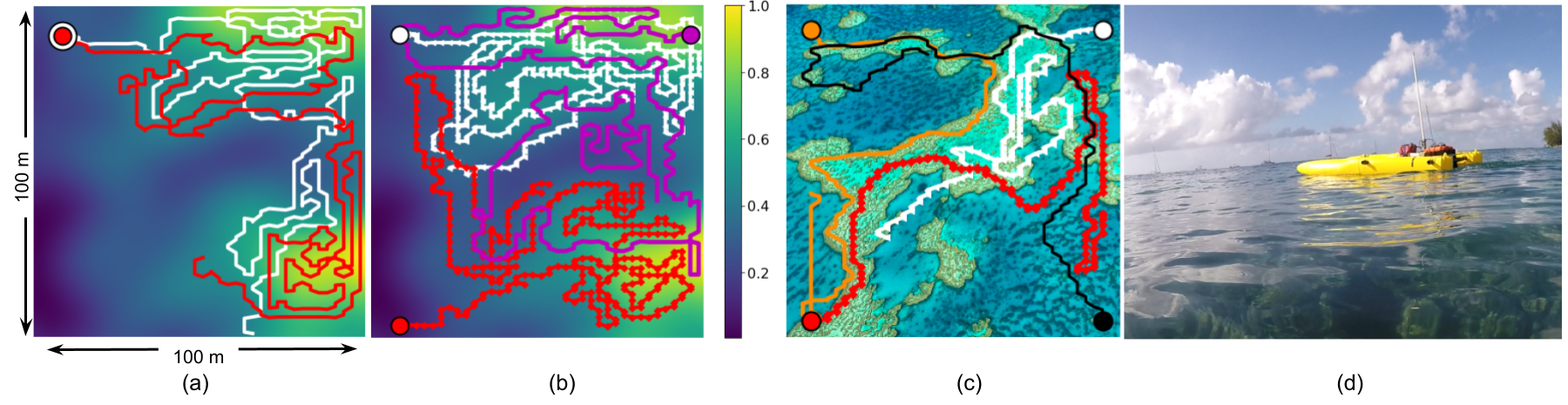}
		\caption{(a) Two-robot and (b) Three-robot sampling paths planned over bathymetry data collected during our field trials in Barbados. (c) Four-robot sampling paths planned over a satellite image of a coral reef. Colored circles represent the start points of the robots. (d) An autonomous surface vehicle (ASV) deployed to sample the bathymetry and visual data over the reefs in Barbados.}
		\vspace{-1.5em}
		\label{fig:result4}
	\end{center}
\end{figure*}

\begin{figure*}
	\begin{center}
		\centering
		\vspace{0.5em}
		\includegraphics[width=\textwidth, height=0.2\textheight]{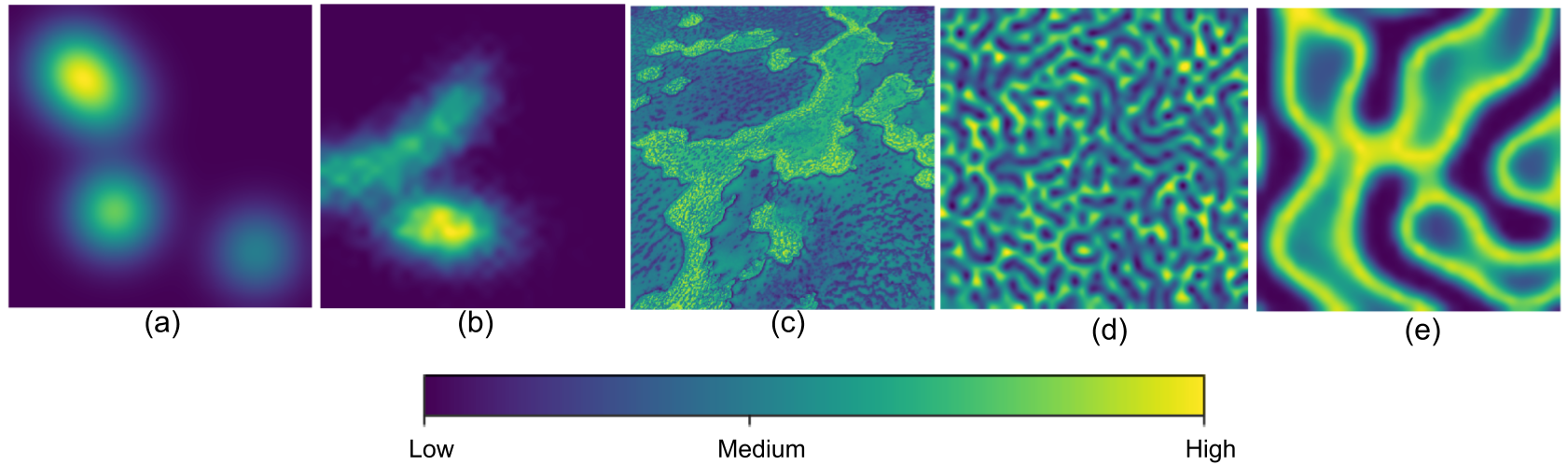}
		\caption{Distributions used for analyzing the multi-robot policy gradient spatial sampling algorithm. The first two distributions ((a) and (b)) are generated synthetically using a mixture of Gaussians. Distribution in (c) is a scoremap generated using real reef data (multi-spectral processing of an aerial image over a reef). (d) is the diffusive spreading pattern simulating the impacts of different nutrients on algae~\cite{dai2014nonlinear}. (e) is a trimmed section of the distribution in (d).}
		\label{fig:synthetic_test}
		\vspace{-1.8em}
	\end{center}
\end{figure*}

We present an empirical evaluation of our sampling technique with statistically significant results. We demonstrate our methodology with sea-floor bathymetry data collected during our field trial as illustrated in Fig.~\ref{fig:result4}. We also evaluate our sampling approach on a set of diverse environments presented in Fig.~\ref{fig:synthetic_test}, some synthetically generated and others commonly occurring in nature. We compare our sampling algorithm to baseline strategies through simulated experiments that utilize models derived from both synthetic and real robot deployment data.

\section{Related Work}\label{sec:related_work}

There is a large literature on robotic data sampling and coverage. Especially in the marine domain, much of this is based on mechanisms that use waypoints of geometric priors with limited dependence on the distribution of incoming observations
~\cite{Shkurti12iros, pizarro, Kemna-2018-986, manjanna2017data}. Many techniques use data regarding boundaries of the environment~\cite{choset1996sensor} and some methods try to minimize odometry error or energy utilization~\cite{sipahioglu2010energy,xu2011optimal}. Our work places emphasis on the density of valuable measurements that can be collected in a limited amount of time.

The design of pragmatic, efficient coordinated multi-robot exploration algorithms to carry out the mission reliably and quickly is a topic of active research. The use of multiple robots instead of a single robot is often suggested to have several advantages and leads to a variety of design trade-offs. Classical algorithmic results related to computational complexity perform poorly in stochastic and time-varying real-world scenarios~\cite{MultirobotRendezvousDudkRoy1997, yan2013survey, higuera2013fair}. Another advantage of robot teams is due to the merging of overlapping information, which can help compensate for sensor uncertainty~\cite{burgard2005coordinated, almadhoun2019survey}.

Robotic sampling requires an efficient strategy to decide where to sample so that there is the highest knowledge gain. Decision making can be regarded as a cognitive process resulting in the selection of one or more actions ({\it i.e.}, a policy) among several alternatives. Two popular approaches for decision making in multi-robot scenarios are centralized and decentralized. A decentralized approach for decision making is scalable, robust, and efficient~\cite{dudek1996taxonomy, vaughan2002lost}. Generally, the problem of distributed multi-robot exploration can be stated as $n$ identical robots set out to explore an unknown area, each robot is equipped with sensing, localization, mapping, and limited-range communication capability~\cite{sheng2006distributed}. Julian~\etal propose an information theoretic approach to distributively control multiple robots equipped with sensors to infer the state of an environment~\cite{julian2012distributed}. In their approach, the robots iteratively estimate the environment state using a sequential Bayesian filter, while continuously moving along the gradient of mutual information to maximize the informativeness of the observations provided by their sensors. However, they assume a connected communication graph throughout the exploration task. Instead, in our approach, we consider a fixed distance communication disk model around the robot~\cite{spanos2005motion, ji2007distributed, zavlanos2008distributed}. This allows for intermittent communication between the robots, which is more realistic in large-scale applications.

Even though multi-robot systems have numerous advantages, they do have additional costs. One such overhead is coordination and communication between the robots. The overall system performance can be directly affected by the quality of coordination and control. Communication, as a means of coordination, enables robots to share position information, state of the environment, sensor measurements, and enable individual robots to learn about the intentions, goals, and actions of other robots. Yan~\etal, in their survey, classified the communication structure based on the information transfer modes, such as explicit and implicit communication~\cite{yan2013survey}. Explicit communication allows direct exchange of information between the robots~\cite{freda20193d, lowe2017multi, salam2019adaptive}, whereas in implicit communication, robots get information about their fellow robots through the environment~\cite{godoy2016implicit}. Explicit communication model enables the robots to work in a team towards completing the task efficiently. In this paper, we explore explicit and asynchronous communication between robots with a constraint on the communication range.

Compared to complete coverage algorithms, adaptive sampling approaches trade off completeness for efficiency. In many cases, even if the process being studied is rapidly varying, subsampling can be effective when the sample points are correctly selected~\cite{venkataramani2000perfect}. When the environmental phenomena being sampled are smoothly varying without any local maxima peaks, non-adaptive strategies are known to perform well~\cite{singh2006active}. However, if the environment contains peaks with high local-variance, adaptive sampling can exploit the clustering phenomena to map the environmental field more accurately than non-adaptive sampling~\cite{manjanna2017data}. In our multi-robot adaptive sampling approach, we explore the possibility of multiple independently trained robots performing the sampling task efficiently with minimal communication between the robots. Our formulation does not encompass the objective of minimizing the communication, but it is a by-product of our sampling approach. We compare the overall sampling performance of the team by varying the communication range of the robots, thus evaluating our approach for scalability in large-scale sampling problems.


\section{Problem Formulation}\label{sec:formulation}

The sampling region is a continuous two-dimensional area of interest $\mathcal{E} \subset \mathbb{R}^2$ with user-defined boundaries. The spatial sampling region is discretized into uniform grid cells, such that the $k^{th}$ robot's position $\robotloc$ can be represented by a pair of integers $\robotloc \in \mathbb{Z}^2$. Each grid cell $(i,j)$ is assigned a prior score value $\rmap(i,j)$ of the data that can be sampled in that cell. We assume that a low-quality initial estimate of the phenomenon being sampled is known either through pilot surveys or satellite data and use this prior to initialize the grid cell scores.

The objective is to maximize the total accumulated score by all $K$ robots, $J_{total}=\sum_{k=1}^{K} J_k$. This in turn can be achieved by maximizing the accumulated score by each robot $J_k$. For brevity, we will use $J$ to indicate each robot's accumulated score. Thus, the goal of an individual robot is to maximize $J$ over a trajectory $\tau$ within a fixed amount of time $T$. We will specify each robot's behavior using a parameterized policy. This is a conditional probability distribution $\pi_{\vec{\theta}}(\vec{s},\vec{a}) = p(\vec{a}|\vec{s};\vec{\theta})$ that maps the current state $\vec{s}$ of the survey to a distribution over possible \emph{actions} $\vec{a}$. Our aim will be to automatically find good parameters $\vec{\theta}$, after which the policy can be deployed without additional training on sampling new environmental fields.

Finding a sequence of actions that maximizes a long-term objective could be done using dynamic programming or other iterative techniques. However, in our formulation, the system state is described using a map containing the per-cell score or value of the physical phenomenon being measured and this scoremap changes as areas are visited by the robot. Hence, the state is composed of the agent's current position and the scoremap containing the per-location value of the phenomenon being measured. As a result of formulating complex states, the state space grows exponentially with the size of the sampling region. Dynamic programming is impracticable with such a large state space --- especially if the time to solve each particular problem is limited. Instead, we turn to methods that directly optimize the policy parameters $\vec{\theta}$ based on (simulated) experiences. To apply these methods, we will first formalize the sampling problem as a Markov Decision Process(MDP).

\subsection{Formalizing Sampling as an MDP}\label{sec:sampling_mdp}

An MDP is a formal definition of a decision problem that is represented as a tuple $(S, A,T(\vec{s}_{t+1}|\vec{s}_t, \vec{a_t}),r(\vec{s}_t, \vec{a}_t),\gamma)$, where $S$ and $A$ are the state and action space, $T$ models transition dynamics based on the current state and action, and $r$ defines the reward for the current state-action pair. $\gamma$ is a discount factor that reduces the desirability of obtaining a reward $t$ time-steps from now rather than in the present by $\gamma^t$. The objective is then to optimize the expected discounted cumulative reward $J=\mathbb{E}_\tau[\sum_{t=0}^{H} \gamma^{t} r_t(s_t,a_t)]$, where $H$ is the optimization horizon.

Our current sampling formulation considers the state $\state$ to include both position of the $k^{th}$ robot $\robotloc$ as well as the map $\rmap$ containing per-location utility value of the phenomenon being measured, $\state=(\robotloc,\rmap)$. To begin with, we consider a 4-connected action space with the options for the robot to move to and scan the cell North, East, South or West of its current location. The action space can be easily expanded or constrained according to the motion constraints of the platform used. We consider a probabilistic choice of actions, that is, an action is chosen according to the policy distribution $\pi_{\vec{\theta}}(\vec{s},\vec{a})$. Once the data in the current robot location is measured, the utility value $\rmap(i,j)$ of the current grid-cell $(i,j)$ is reduced to $0$. The discounted reward function is defined as $\gamma^t \rmap(\robotloc)$, with the discount factor $0\le \gamma \le 1$ encouraging the robot to sample cells with high scores in early time steps $t$. Discount factor $\gamma$ with a value closer to $0$ encourages myopic trajectories, whereas a $\gamma$ closer to $1$ pushes towards achieving trajectories that are farsighted (meaning paths that achieve higher accumulated rewards)

\section{Policy Gradient for Sampling}\label{sec:pg}

The complex and large state space required to formulate the sampling problem makes it unrealistic to apply dynamic programming or other iterative techniques for robotic sampling problems. Even though we have used iterative methods to solve similar problems, they do not scale well with the increase in the size of the sampling region~\cite{manjanna2017data}. Hence, we make use of  policy gradient search that directly optimizes the policy parameters $\vec{\theta}$ based on (simulated) experiences.

Policy gradient methods use gradient ascent for maximizing the expected return $J_\theta$. The gradient of the expected return ($\nabla_\theta J_\theta$) guides the direction of the parameter ($\theta$) update. The policy gradient update is given by,

\begin{equation}
\label{eqch6:param_update}
\theta_{k+1} = \theta_k + \eta \nabla_\theta J_\theta,
\end{equation}
where $\eta$ is the learning rate. The policy gradient is given by,
\begin{equation}
\label{eqch6:update}
\nabla_\theta J_\theta = \int_\tau \nabla_\theta p_\theta(\tau) R(\tau) d\tau,
\end{equation}
where $R(\tau)$ is the reward obtained by following the trajectory $\tau$. One of the effective methods to estimate the gradient $\nabla_\theta J_\theta$ is to make use of the \emph{likelihood-ratio} trick that is given by the identity $\nabla p_\theta(y) = p_\theta(y) \nabla log p_\theta(y)$. Applying this likelihood-ratio trick to the policy gradient in Eq.~\ref{eqch6:update}, decomposing $\nabla_\theta log p_\theta(\tau)$ into single time steps and applying the logarithm yields,
\begin{equation}
\label{eqch6:logp}
\nabla_\theta log p_\theta(\tau) = \sum_{t=0}^{H-1} \nabla_\theta log \pi_\theta(a_t | s_t,t)
\end{equation}

One of the first policy gradient algorithms, REINFORCE Algorithm~\cite{williams1992simple}, applies Eq.~\ref{eqch6:logp} and uses a variance-reducing baseline $b$ to define the policy gradient as,
\begin{equation}
\label{eqch6:pg_reinforce}
\nabla_\theta J_\theta = \mathbb{E}_{p_\theta(\tau)} \left[\sum_{t=0}^{H-1} \nabla_\theta log \pi_\theta(a_t | s_t,t) (R(\tau)-b)\right]
\end{equation}

This expression, however, depends on the correlation between actions and previous rewards. These terms are zero in expectation, but can often induce additional variance. Ignoring these terms yields lower-variance updates. G(PO)MDP algorithm~\cite{greensmith2004variance} and Policy Gradient Theorem (PGT) algorithm~\cite{sutton2000policy} propose to reduce the variance of policy gradient estimates by using the observation that rewards from the past do not depend on actions in the future. Accordingly, the policy gradient is given by,
\vspace{-0.5em}
\begin{multline}
\label{eqch6:pg_gpomdp}
\nabla_\theta J_\theta = \frac{1}{m}\sum_{i=1}^m\sum_{t=0}^{H-1}\nabla_\theta \log \pi_\theta(a_t^{(i)}|s_t^{(i)}) \\ \left(\sum_{j=t}^{H-1}r(s_j^{(i)},a_j^{(i)}) - b(s_t^{(i)})\right).
\end{multline}

The expectation over $p_\theta(\tau)$ is approximated by summing over all the sampled trajectories ($\tau^{(i)} = (s^{(i)}_0, a^{(i)}_0, \\s^{(i)}_1, a^{(i)}_1, ...)$). The gradient is based on $m$ sampled trajectories from the system, with $\vec{s}_j^{(i)}$ the state at the $j^{\text{th}}$ time-step of the $i^{\text{th}}$ sampled roll-outs. Furthermore, $b$ is a variance-reducing baseline. In our experiments, we use the observed average reward as baseline ($b = \mathbb{E}[R(\tau)] \approx \frac{1}{m}\sum_{i=1}^m R(\tau^{(i)})$).

\subsection{Policy Design}\label{sec:policy_design}

An effective approach to define stochastic policies over a set of deterministic actions is the use of Gibbs distribution of a linear combination of features as a policy (also referred to as Boltzmann exploration of the softmax policy). We consider a commonly used linear Gibbs softmax policy parametrization~\cite{sutton2000policy} given by,
\begin{equation}
\label{eq:policy}
\pi(\mathbf{s},\mathbf{a}) = \frac{e^{\theta^T\phi_{sa}}}{\sum_be^{\theta^T\phi_{sb}}},\hspace{1cm}\forall s \in S; a,b \in A,
\end{equation}
where $\phi_{sa}$ is an $k$-dimensional feature vector characterizing a state-action pair ($s,a$) and $\theta$ is an $k$-dimensional parameter vector. This is a commonly used policy in reinforcement learning approaches. 

The final feature vector $\phi_{sa}$ is formed by concatenating a vector $\phi'_{s} \delta_{a a'}$ for every action $a' \in [North, East, \\South, West]$, where $\phi'_{s} \subset \mathbb{R}^k$ is a feature representation of the state space, and $\delta_{a a'}$ is the Kronecker delta. Thus, the final feature vector has $4 \times k$ entries, $75\%$ of which corresponding to non-chosen actions will be $0$ at any one time step.

\begin{figure}[h]
	\begin{center}
		\centerline{\includegraphics[width=\columnwidth]{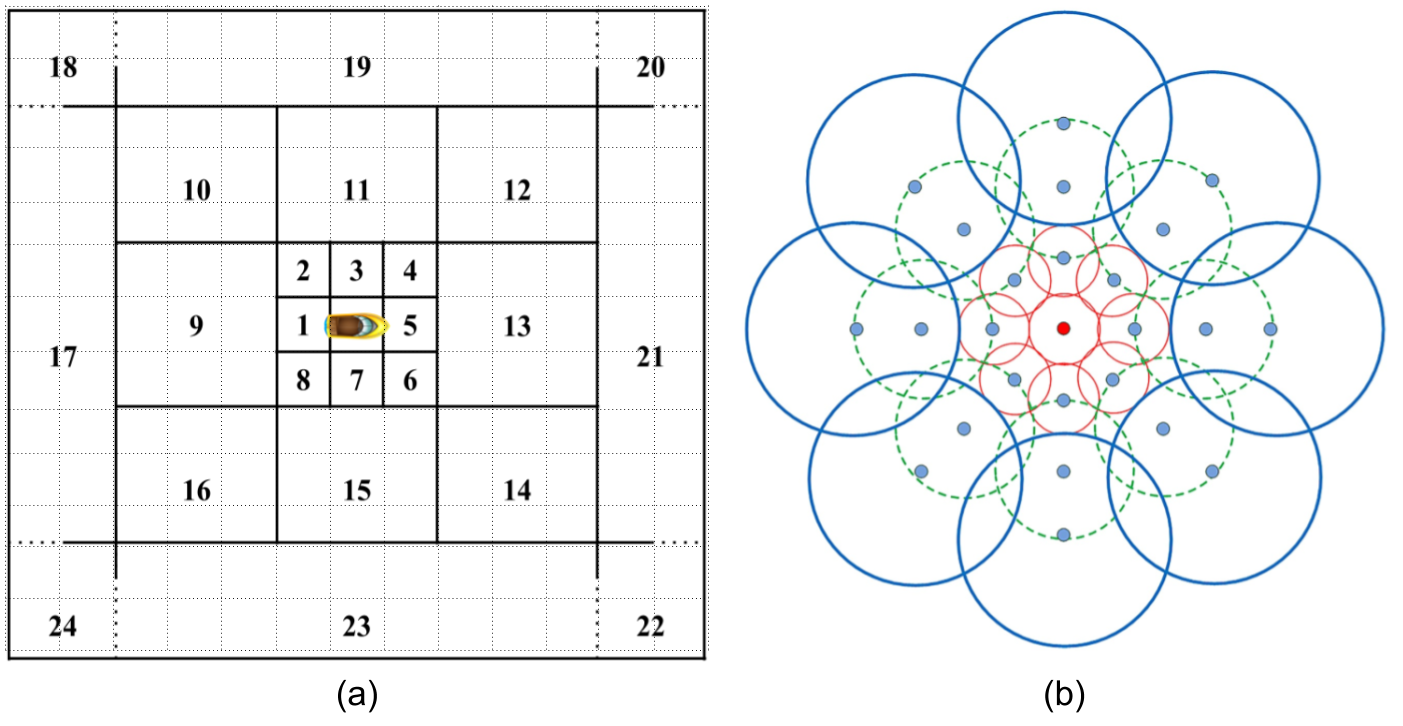}}
		\caption{(a) Multiresolution aggregation of feature space. (b) DAISY: local image descriptors that inspired our feature aggregation~\cite{tola2009daisy}.}
		\label{fig:multi-features}
		\vspace{-4.5em}
	\end{center}
\end{figure}

\subsection{Feature Design}\label{sec:feature_design}

In our previous work~\cite{manjanna2018policy}, we proposed and evaluated multiple feature aggregation designs. Our empirical analysis showed that having a multiresolution aggregation of the feature space resulted in achieving better discounted rewards. In multiresolution feature aggregation, feature cells grow in size along with the distance from robot location as depicted in Fig.~\ref{fig:multi-features}(a). Thus, areas close to the robot are represented with high resolution and areas further from the robot are represented in lower resolution. We draw our inspiration for this feature design from a popular local image descriptor called DAISY descriptor~\cite{tola2009daisy} presented in Fig.~\ref{fig:multi-features}(b). The intuition behind this feature design is that the location of nearby reward values is important to know exactly, while the location of faraway rewards can be represented more coarsely. The multiresolution feature design is also suitable for bigger worlds as it scales logarithmically with the size of the world.

\section{Sampling with Multi-robot Teams}

We propose a decentralized sampling approach where each robot in a team performs an informed survey using a policy-gradient-based sampling strategy as outlined above. As mentioned earlier, we do not formulate the multi-robot scenario as a learning problem. Instead, we train each robot as an independent agent. 

\revision{One way to achieve collaborative sampling between multiple independent robots is by assigning a reward function for the robot $r_k$ that is proportional to the distance of grid location $l_i$ from other $n-1$ coworking robots and inversely proportional to the distance of grid location $l_i$ from its current location $l_{r_k}$. Thus encouraging robot $k$ to visit locations that are close to the $l_{r_k}$ and farther from other co-working agents. This results in dividing the task spatially so that each robot has its own subregion to sample.}

\begin{equation} \label{reward}
R(l_{r_k},l_i) \propto \frac{\displaystyle\sum_{\substack{j \in (1,2,..n) , j \ne k}} distance(l_{r_j},l_i)}{distance(l_k,l_i)}.
\end{equation}

\revision{However, performing such weighted reward adjustments can compromise the resilience and robustness of the multi-robot team. Overall, it will degrade the sampling performance as the robots can and do get stuck within their own self-defined region of space. Instead, in our proposed approach, the robots communicate only their visited locations with the neighboring robots within their communication range. This information is used by each robot to transition to the state where the rewards from all the visited locations are removed. In this work, we explore how different communication ranges impact the overall sampling performance of the robot team. For our empirical analysis, we run experiments on the distributions presented in Fig.~\ref{fig:synthetic_test}.}

\begin{figure*}
	\centering
	\subfigure[]{ \includegraphics[width=0.205\textwidth]{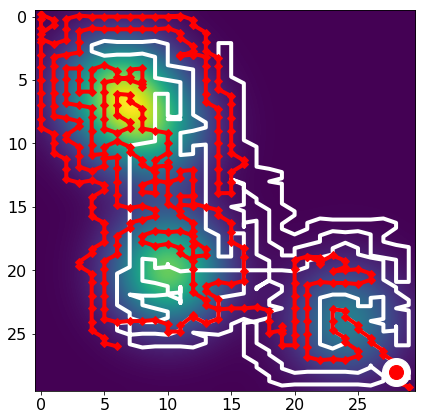} \label{fig:gauss_path}}\hspace{-1em}
	\subfigure[]{ \includegraphics[width=0.39\textwidth, height=0.16\textheight]{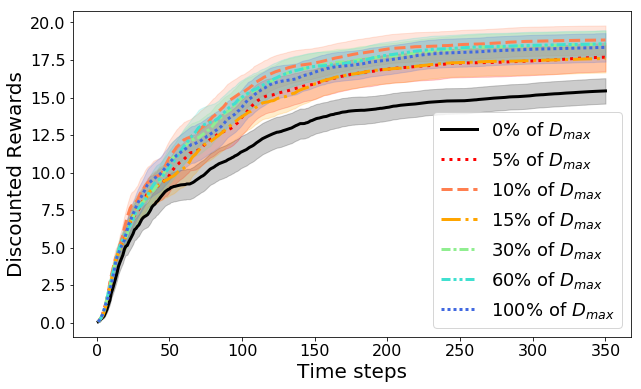} \label{fig:gauss_dis}}\hspace{-1em}
	\subfigure[]{ \includegraphics[width=0.39\textwidth, height=0.163\textheight]{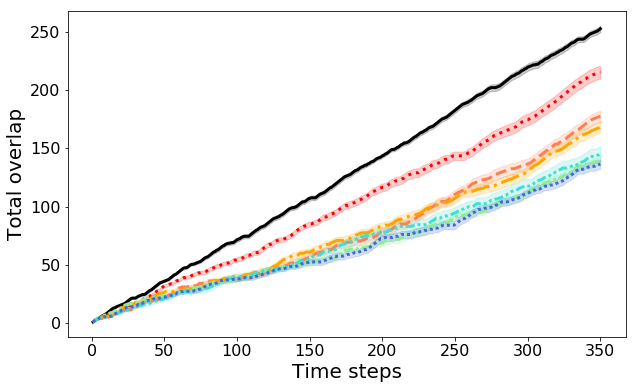} \label{fig:gauss_overlap}}\vspace{-5pt}
	\subfigure[]{ 
	\includegraphics[width=0.205\textwidth]{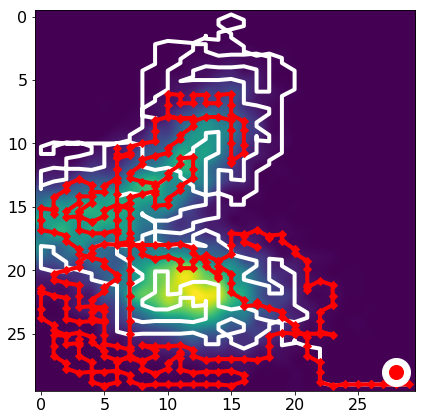} \label{fig:c_path}}\hspace{-1em}
	\subfigure[]{ \includegraphics[width=0.39\textwidth]{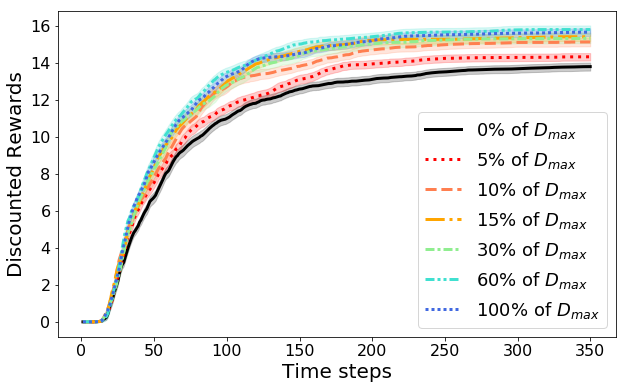} \label{fig:c_dis}}\hspace{-1em}
	\subfigure[]{ \includegraphics[width=0.39\textwidth]{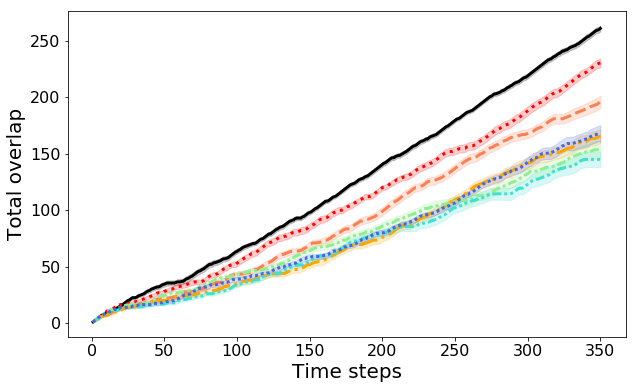} \label{fig:c_overlap}}\vspace{-5pt}
	\subfigure[]{ \includegraphics[width=0.205\textwidth]{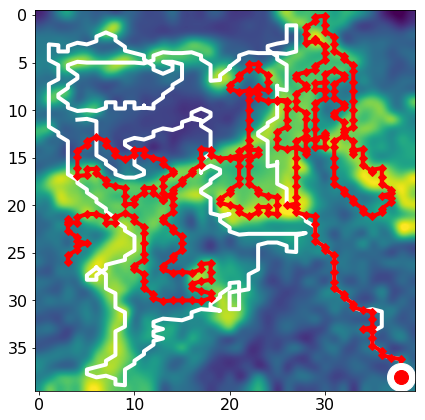} \label{fig:reef_path}}\hspace{-1em}
	\subfigure[]{ \includegraphics[width=0.39\textwidth]{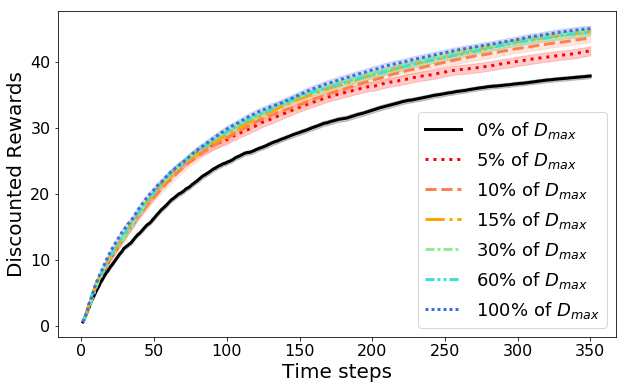} \label{fig:reef_dis}}\hspace{-1em}
	\subfigure[]{ \includegraphics[width=0.39\textwidth]{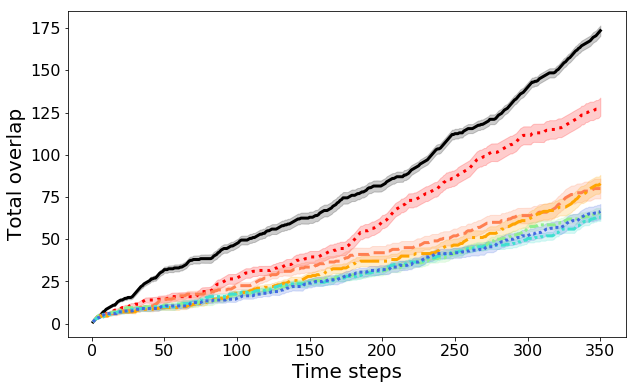} \label{fig:reef_overlap}}\vspace{-5pt}
	\subfigure[]{ \includegraphics[width=0.205\textwidth]{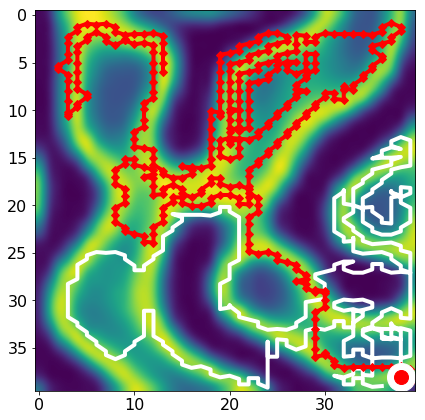} \label{fig:diffusion_path}}\hspace{-1em}
	\subfigure[]{ \includegraphics[width=0.39\textwidth]{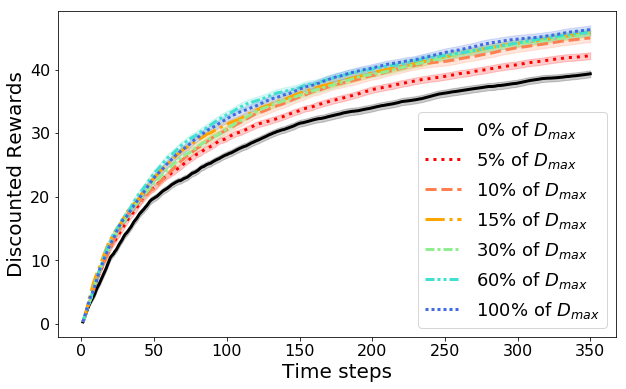} \label{fig:diffusion_dis}}\hspace{-1em}
	\subfigure[]{ \includegraphics[width=0.39\textwidth]{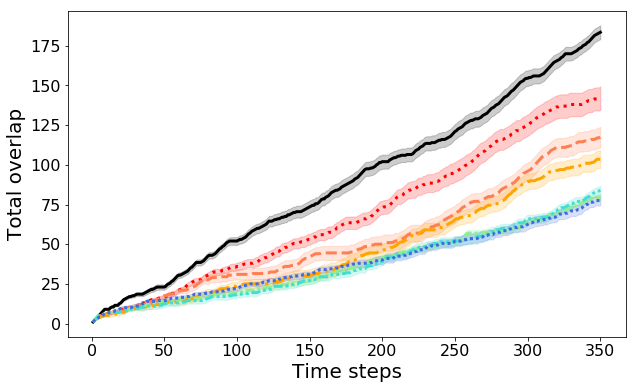} \label{fig:diffusion_overlap}}\vspace{-5pt}
	\subfigure[]{ 
	\includegraphics[width=0.205\textwidth]{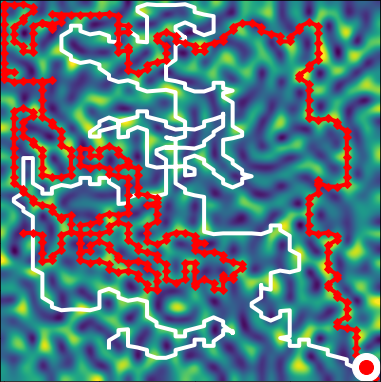} \label{fig:diffusion_full_path}}\hspace{-1em}
	\subfigure[]{ \includegraphics[width=0.39\textwidth]{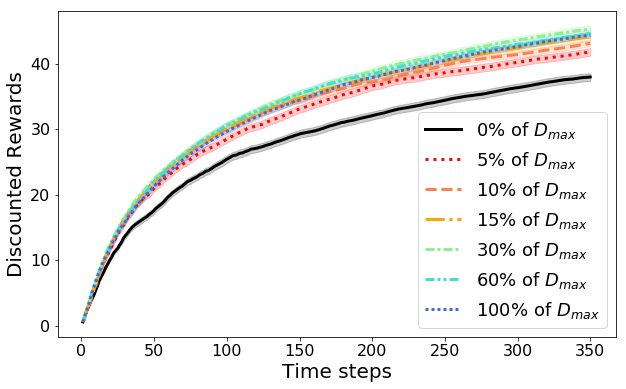} \label{fig:diffusion_full_dis}}\hspace{-1em}
	\subfigure[]{ \includegraphics[width=0.39\textwidth]{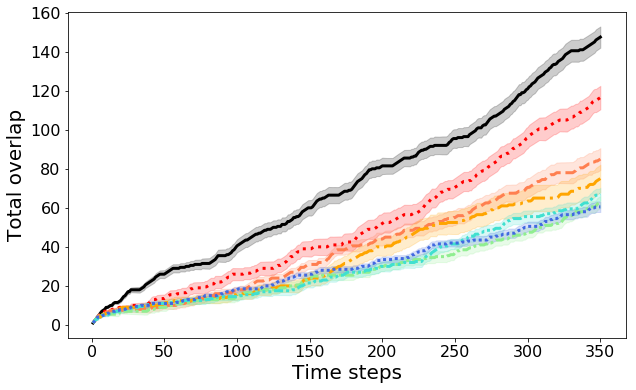} \label{fig:diffusion_full_overlap}}\vspace{-5pt}
	\caption{Sampling performance and illustrative paths of two robots operating with distance constrained communication. The shaded regions around the median line illustrate the standard error over $20$ trials for each communication range.}
	\vspace{-1em}
	\label{fig:result1}
\end{figure*}

\subsection{Distance constrained communication}

We explore the ability of a multi-robot team to sample efficiently with a distance-limited communication between the robots. Every robot can communicate only its current state with the neighboring robots that are within the communication range. This state information is then used by neighboring robots to update their sampling plan accordingly. 

This setup with distance constrained communication imitates the real-world scenario in any outdoor deployment environment. Especially on the surface of water, the communication signals attenuate faster~\cite{coelho2018experimental} making it infeasible to exchange messages between robots that are far apart. We define the distance constraints as a percentage of the maximum possible distance ($D_{max}$) in the bounded survey region. Thus, $0\%$ represents no communication and $100\%$ represents complete communication throughout the survey region.

We note that robots can choose the same overlapping path when there is no communication between them under this strategy. Sometimes if the robots are synchronized in sampling, they will end up following the same path even after communicating on every step as they have knowledge only about the current step. Moreover, having such frequent communication is expensive and infeasible in the real world. Our previous observations show that a communication after the collection of every $10$ to $30$ samples would fetch good rewards.

\section{Experimental Results}

\begin{figure*}
	\centering
	\hspace{-1em}
	\subfigure[]{ \includegraphics[width=0.48\textwidth, ]{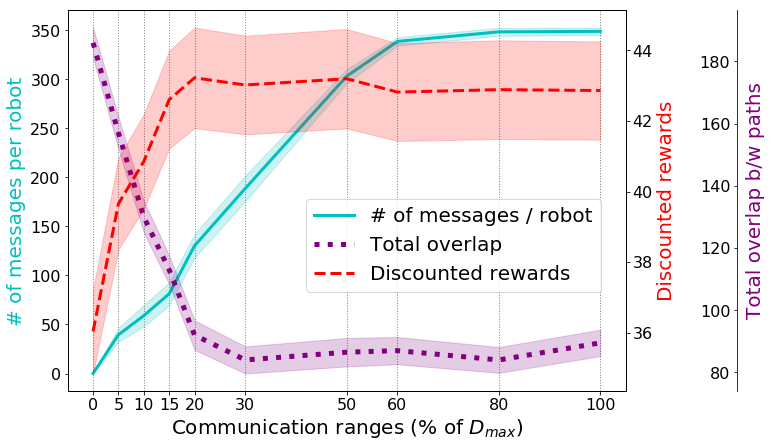} \label{fig:comms_overlap}}\hspace{3em}
	\subfigure[]{ \includegraphics[width=0.4\textwidth]{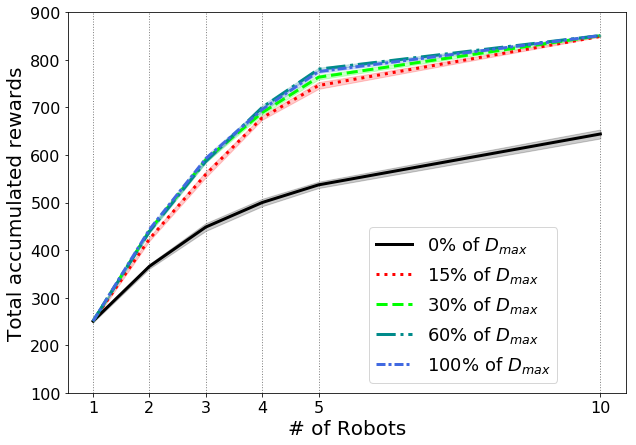} \label{fig:num_robots}}
	\subfigure[]{ \includegraphics[width=0.8\textwidth, ]{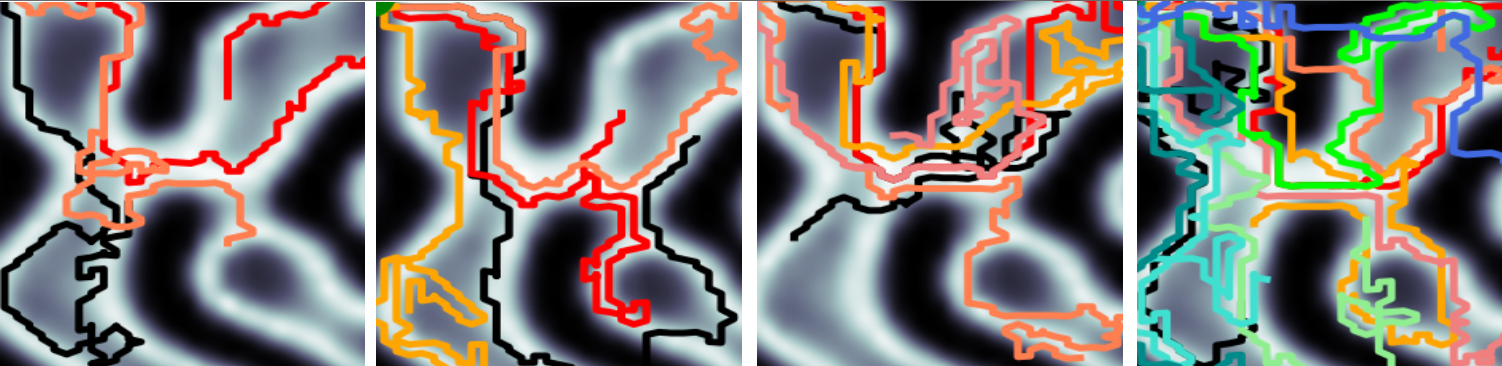} 
	\label{fig:multi-robot}}
	\caption{(a) Accumulated rewards as the number of robots is increased. The shaded area represents the standard error over $20$ trials. (b) Sampling performance metrics averaged over all the $5$ distributions. These results illustrate higher sampling performance achieved even with minimal communication. The shaded area represents the standard error over $100$ trials. (c) Paths for varying team size ($2, 4, 5,$ and $10$) on an example static spatial field.}
	\vspace{-1.5em}
	\label{fig:result2}
\end{figure*}

\subsection{Setup}
We train all our robots on a set of distributions generated by a mixture of Gaussians. All robots have identical learnt parameters, thus are independently capable of collecting good samples in the data collection task. A diverse set of five distributions is used to simulate the physical phenomenon of interest. Two of these distributions are synthetically generated using a mixture of Gaussians (Fig.~\ref{fig:synthetic_test}(a) and ~\ref{fig:synthetic_test}(b)). The next distribution in Fig.~\ref{fig:synthetic_test}(c) is a scoremap generated by multispectral image processing on an aerial image of the coral reef. Last two distributions in Fig.~\ref{fig:synthetic_test}(d) and ~\ref{fig:synthetic_test}(e) are generated using the model for a diffusive spreading pattern simulating the impacts of different nutrients on algae~\cite{dai2014nonlinear}. Over the multiple trials, we combine different starting states of the robots (such as all robots starting at a fixed location, or randomly chosen location, or every robot starting at opposite corners). We experimented with $10$ communication ranges. In this section, we only present a small set of our experimental results to achieve better readability and clarity in the plots. We apply Savitzky–Golay filter~\cite{savitzky1964smoothing} to smoothen the planned paths. This kind of path smoothing is used only if the paths are planned offline. 
Many applications in the domain of environmental monitoring with autonomous vehicles have a need to sample in information rich locations at the earliest~\cite{low2008adaptive}. This is true when robotic vehicles have a limited endurance to carry out the survey and the environmental processes are quasistatic. Hence, we measure the total accumulated discounted rewards as one of the performance metrics. \revision{In our experiments we use hyperbolic discounting for training as it encourages a non-myopic reward collection by the agent.} Two or more robots surveying a region to collect samples are expected to cover different regions so that the resources are utilized efficiently. Hence, we measure the overlap between the robot paths as the second metric for sampling performance. To achieve an efficient spatial sampling, we want the planned paths to achieve as little overlap as possible and as high discounted rewards as possible.

\begin{figure*}
	\centering
	\subfigure[]{ \includegraphics[width=0.2\textwidth]{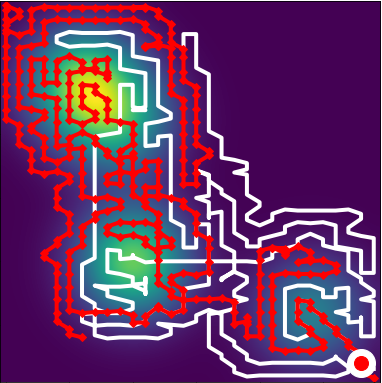} \label{fig:gauss_our}}\hspace{-0.8em}
	\subfigure[]{ \includegraphics[width=0.2\textwidth]{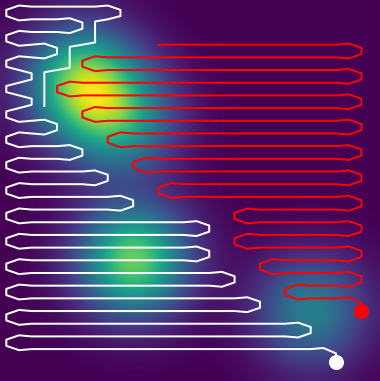} \label{fig:gauss_darp}}\hspace{-0.8em}
	\subfigure[]{ \includegraphics[width=0.2\textwidth]{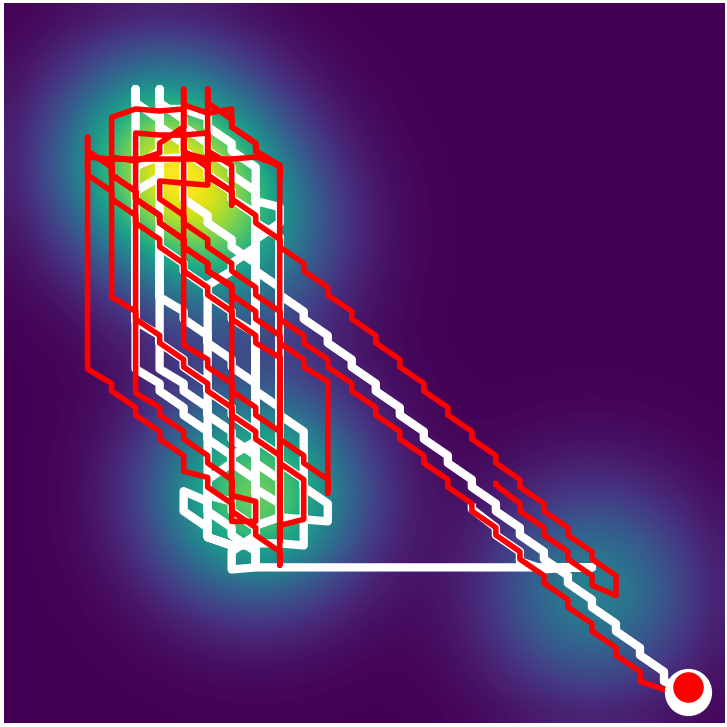} \label{fig:gauss_maxima}}\hspace{-0.8em}
	\subfigure[]{ \includegraphics[width=0.37\textwidth]{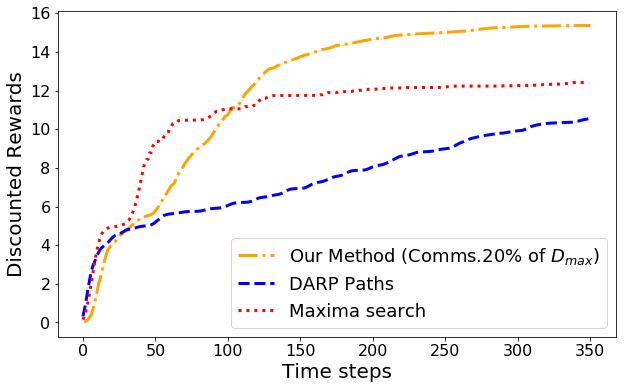} 
	\label{fig:gauss_disc}}\vspace{-5pt}
	\subfigure[]{ \includegraphics[width=0.2\textwidth]{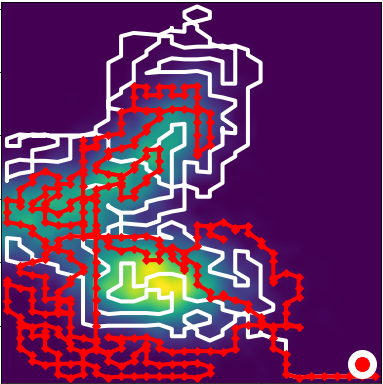} \label{fig:c_our}}\hspace{-0.8em}
	\subfigure[]{ \includegraphics[width=0.2\textwidth]{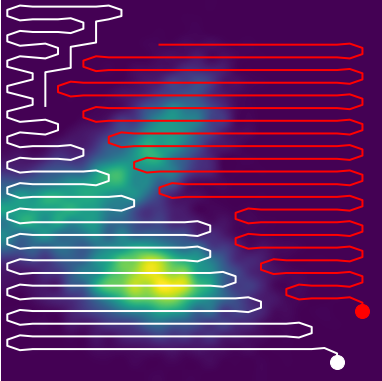} \label{fig:c_darp}}\hspace{-0.8em}
	\subfigure[]{ \includegraphics[width=0.2\textwidth]{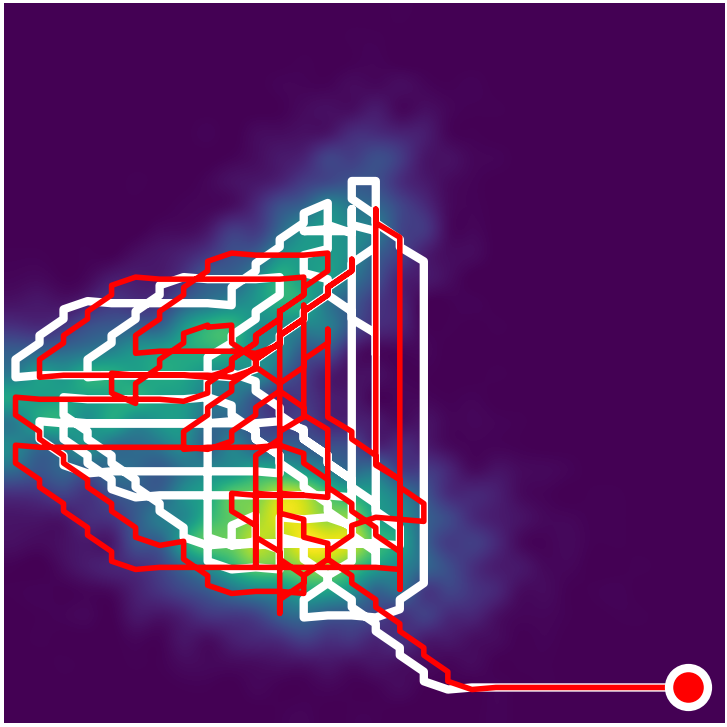} \label{fig:c_maxima}}\hspace{-0.8em}
	\subfigure[]{ \includegraphics[width=0.37\textwidth]{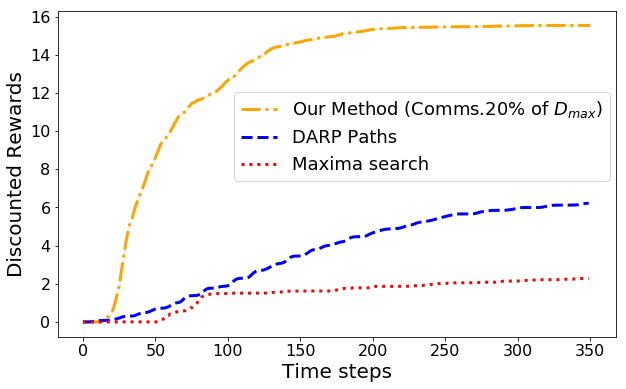} \label{fig:c_disc}}\vspace{-5pt}
	\subfigure[]{ \includegraphics[width=0.2\textwidth]{Diffusion_full_path.png} \label{fig:diffusion_full_our}}\hspace{-0.8em}
	\subfigure[]{ \includegraphics[width=0.2\textwidth]{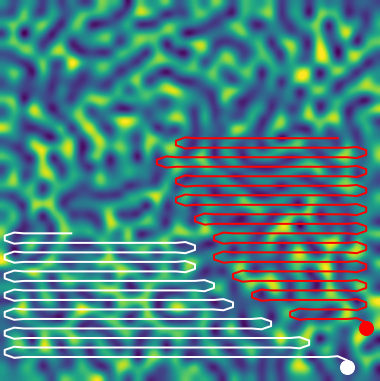} \label{fig:diffusion_full_darp}}\hspace{-0.8em}
	\subfigure[]{ \includegraphics[width=0.2\textwidth]{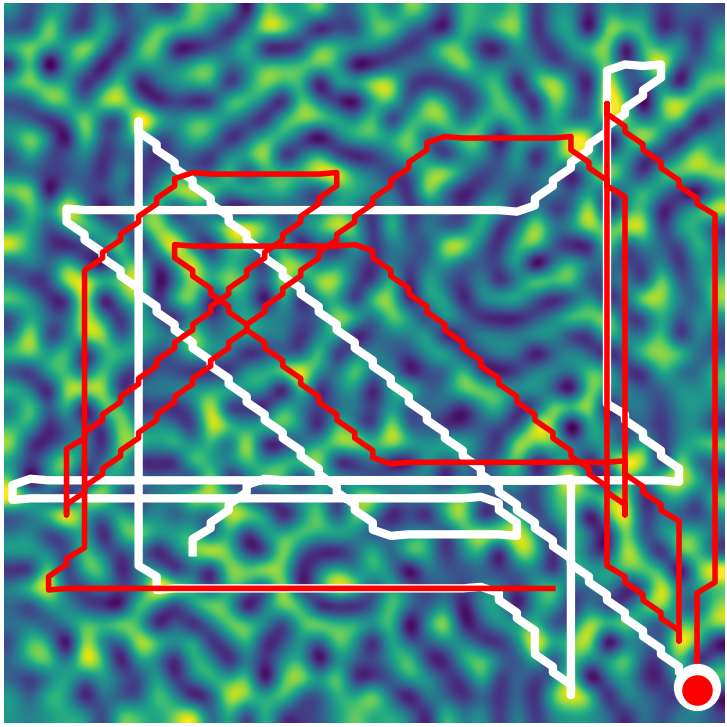} \label{fig:diffusion_full_maxima}}\hspace{-0.8em}
	\subfigure[]{ \includegraphics[width=0.37\textwidth]{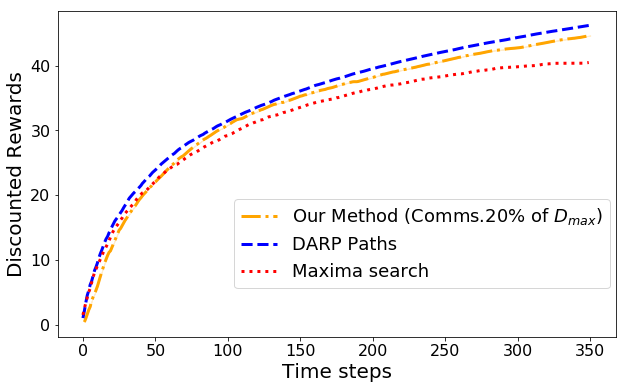} \label{fig:diffusion_full_disc}}\vspace{-5pt}
	\subfigure[]{ \includegraphics[width=0.2\textwidth]{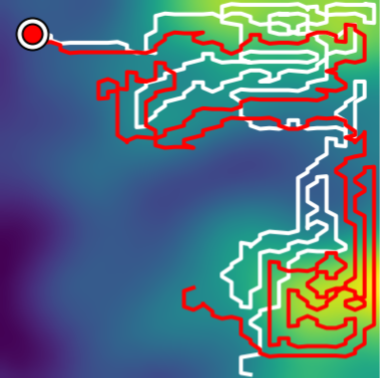} \label{fig:real_path}}\hspace{-0.8em}
	\subfigure[]{ \includegraphics[width=0.2\textwidth]{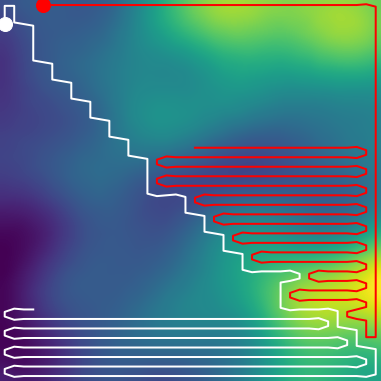} \label{fig:real_dis}}\hspace{-0.8em}
	\subfigure[]{ \includegraphics[width=0.2\textwidth]{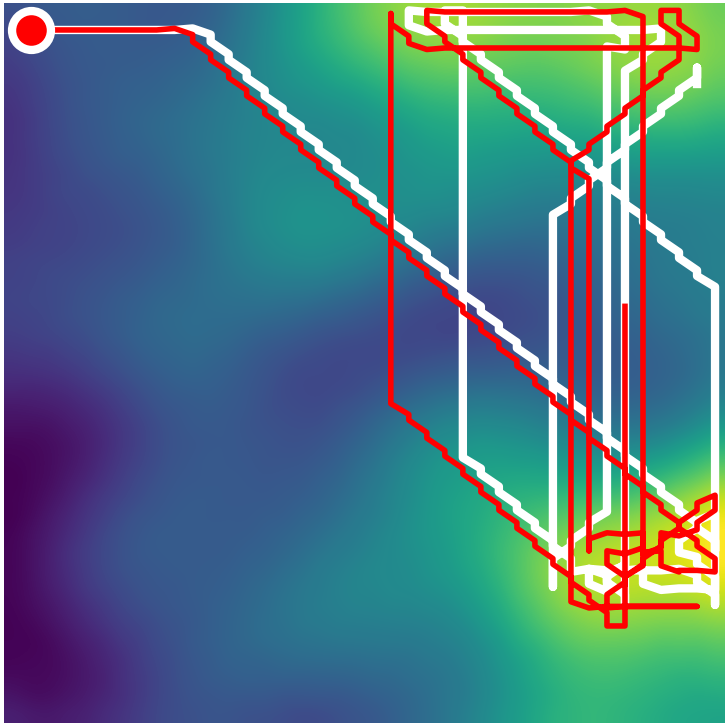} \label{fig:real_maxima}}\hspace{-0.8em}
	\subfigure[]{ \includegraphics[width=0.37\textwidth]{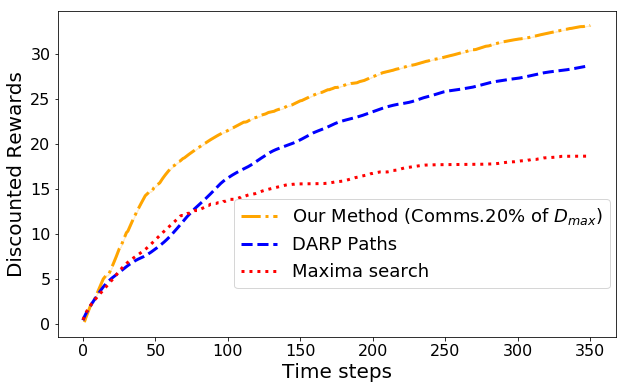} \label{fig:real_disc}}\vspace{-5pt}
	\caption{Comparing the performance of our sampling approach with two baseline techniques. Illustrative paths of two robots following our algorithm and the baseline algorithms.}
	\vspace{-1.5em}
	\label{fig:result3}
\end{figure*}

\subsection{Results and Discussion}

Fig.~\ref{fig:result1} demonstrates the results from our experiments involving two robots starting their sampling path from the same starting point. The first column in Fig.~\ref{fig:result1} illustrates the example paths generated by two robots with full communication. The second and the third columns in Fig.~\ref{fig:result1} present the discounted rewards and the path overlap, respectively. The shaded area specifies the standard error of the mean over $20$ trials for each of the $7$ communication ranges. These results illustrate that, irrespective of the underlying distribution, a communication range between $10\%$ and $30\%$ of the $D_{max}$ can achieve good sampling with minimal message exchange. To confirm this hypothesis, we further analyze the results by averaging over all $5$ spatial distributions.

Fig.~\ref{fig:comms_overlap} presents the averaged results over all spatial distributions plotted against the communication range on the x-axis. This plot indicates a significant improvement in the reward collection and a significant decrease in the path overlap once the communication range is over $15\%$ of the maximum possible distance $D_{max}$. Another observation from Fig.~\ref{fig:comms_overlap} is that the robots are achieving good sampling performance with minimal message exchange between them. This resulting behavior of increased sampling performance with minimal communication is a desired feature for a decentralized multi-robot sampling approach.

We evaluate the scalability of our approach by increasing the number of robots involved in sampling (Fig.~\ref{fig:multi-robot}). The plots presented in Fig.~\ref{fig:num_robots} illustrate the change in total rewards collected by the robotic team as the number of deployed robots increases (x-axis). It can be observed that the curves for the total rewards collected plateaued with the increase in the number of robots. This is because of the bounded sampling region where all robots are fenced according to our formulation of the problem.

\begin{figure*}
	\centering
	\subfigure[]{ \includegraphics[width=0.24\textwidth]{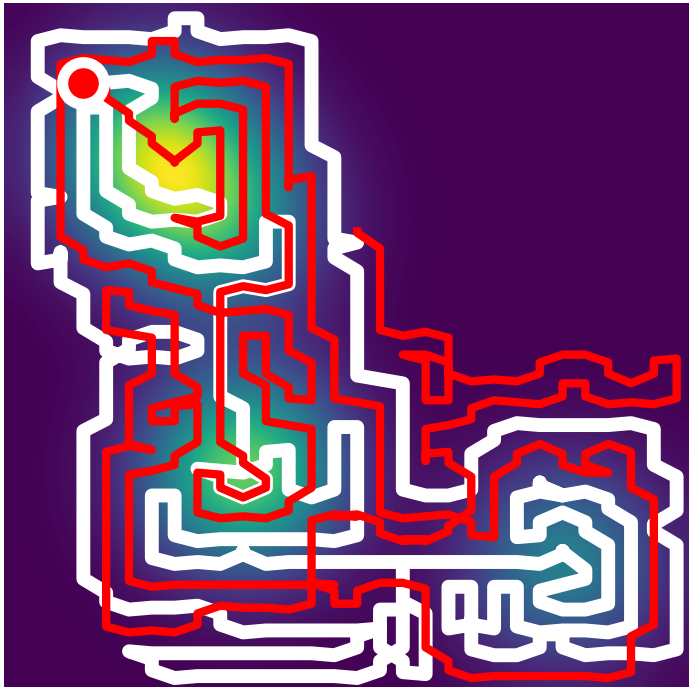} \label{fig:no_fail}}\hspace{-0.9em}
	\subfigure[]{ \includegraphics[width=0.24\textwidth, ]{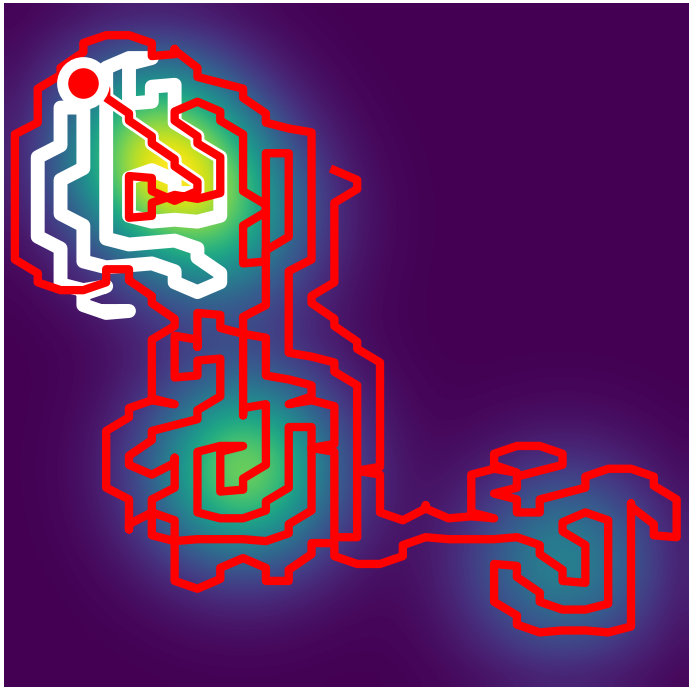} \label{fig:fail_25}}
	\subfigure[]{ \includegraphics[width=0.48\textwidth, ]{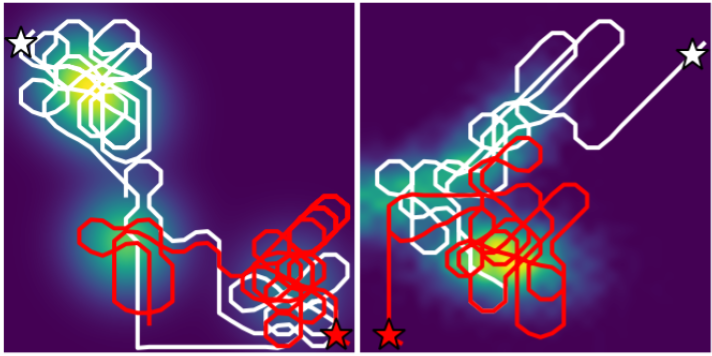} 
	\label{fig:action_constraints}}
	\caption{(a) Sampling paths of the robots when there are no failures. (b) Sampling paths of the robots when when one of the robots fail after first $75$ time steps. (c) Paths planned for two robots sampling with a constrained action space. Colored stars represent the start points of the robots.}
	\label{fig:result5}
\end{figure*}

We also evaluated the robot sampling technique on real bathymetry data collected using two ASVs (Fig.~\ref{fig:result4}(d)) during our field trip in Barbados in 2019. Fig.~\ref{fig:result4}(a) and~\ref{fig:result4}(b) present the paths of multiple robots sampling from these depth maps, where larger rewards are associated with shallower regions, thus inducing the robots to cover relatively shallow regions first. Fig.~\ref{fig:result4}(c) demonstrates the path planning performed on an aerial image of the coral reef. We demonstrate the resulting paths when a varying number of robots start at different locations, as illustrated in Fig.~\ref{fig:result4}. These results establish the applicability of our distributed sampling technique for real-world data sampling scenarios.

\revision{Experiments in Fig.~\ref{fig:result4}(a) present the sampling of bathymetry data of the seafloor using two autonomous surface vehicles. Water in its pure form is an insulator, but as found in its natural state, it contains dissolved salts and other matter which makes it a partial conductor. The higher its conductivity, the greater the attenuation of radio signals which pass through it~\cite{vk5br1987underwater}. In a recent study~\cite{dala2021design}, a Long Range (LoRa) communication antenna was found to achieve a communication range between $80m$ to $160m$ depending on whether that antenna was buffering or not. Using our proposed algorithm, a region of $1km \times 1km$ can be efficiently sampled by two robots with a communication range of $140m$, which is a realistic range on the surface of water to achieve reliable communication connection.}

\vspace{-1.5em}

\subsection{Performance Comparison with baseline techniques}\label{sec:comparison}

We compare our proposed algorithm to two baseline sampling strategies through simulated experiments that utilize models derived from both synthetic and real robot deployment data. The first comparison algorithm is a multi-robot coverage path planning technique called DARP algorithm (Divide Area based on the robot’s initial Positions)~\cite{kapoutsis2017darp}. DARP algorithm divides the region into a number of equal areas each corresponding to a specific robot, such that a complete coverage is guaranteed, the coverage path is of minimum cost and non-backtracking. The second comparison algorithm is a multi-robot adaptation of the maxima search algorithm~\cite{meghjani2016multi}. Each robot sequentially selects the current maxima in the given spatial field and plans a path to that location using A-star path planner. Once the value at a given location is collected, it is reduced to $0$ as in our proposed approach. The first three columns of Fig.~\ref{fig:result3} present the illustrations of fixed horizon paths executed by the two robots using our proposed approach, the DARP algorithm, and the multi-robot maxima search algorithm, respectively. The comparison of discounted rewards achieved by all three techniques is presented in the fourth column of Fig.~\ref{fig:result3}.

We evaluate our proposed approach by comparing the performance in terms of total accumulated rewards and the total discounted rewards. We found that our proposed policy gradient based sampling technique performs better than the other two baseline approaches by collecting high rewarding samples in the early phase of their surveys. As illustrated in Fig.~\ref{fig:gauss_disc} and \ref{fig:c_disc}, our approach performs significantly better for sampling fields with uneven hotspot distribution. If the process of interest is uniformly distributed as in Fig.~\ref{fig:diffusion_full_our}, our approach performs comparably with the complete coverage approach (Fig.~\ref{fig:diffusion_full_disc}). Many environmental processes have a nonuniform spatial distribution as presented in the depthmaps of the coral reefs presented in Fig.~\ref{fig:real_path} and our proposed approach was able to collect high rewarding samples at the early phase of the survey as depicted by the results in Fig.~\ref{fig:real_disc}.

\subsection{Additional properties}\label{sec:extensions}

\revision{In this section, we present additional properties and potential extensions to the proposed planning algorithm. The results in Fig.~\ref{fig:fail_25} illustrate an example of the robustness of our multi-robot sampling system. As shown in Fig.~\ref{fig:no_fail}, when there are no failures, both the robots share and cover the hotspot regions. The robots are all individually trained to perform efficient sampling of a spatial field. Hence, even when one of the robots fails during a survey, the other robot samples from the hotspot regions and achieves a good overall performance for the team as depicted in Fig.~\ref{fig:fail_25}. These preliminary results display a fault-tolerance nature of our multi-robot sampling strategy and we would like to further investigate this behavior in the near future.}

\revision{In Section~\ref{sec:sampling_mdp}, we described that the problem formulation and our solution can be easily extended to any kind of action space. The preliminary results validating these claims are illustrated in Fig.~\ref{fig:action_constraints}. Here, we present an example of two robots operating in constrained action space with a suite of only three ``forward moving''  actions available in any state: turn $45^{\circ}$ left, go straight, and turn $45^{\circ}$ right. All these actions are in relation to the current heading of the robot and hence the heading should be included into our state space to plan for this action space. We are working on extending our planner to plan over action spaces designed using motion primitives that are specific to a robotic platform.}

\section{Conclusions and Future Directions}

In this paper, we presented a distributed path planning approach to generate an efficient sampling path for a team of robots. Our objective is to have paths that are highly rewarding in terms of information gain and have minimal overlap. We achieve higher rewards by using data-driven methods to determine the best policy and attain minimal overlap through communication between the robots. Instead of formulating the multi-robot scenario as a learning problem, each robot is trained independently and put together to perform a sampling task. Our analysis of the coordinated sampling behavior shows that the robot team was able to efficiently collect high rewards with a minimal exchange of messages between the robots. 

Our results demonstrate that, irrespective of the underlying distribution that is being sampled, our approach is able to collect high rewards with a limited communication range of $10\%$ to $20\%$ of the $D_{max}$. These outcomes are significant when sampling in large-scale environments where a complete communication between the robots is unrealistic. We also empirically analyzed the scalability and adaptability of our approach with changing the number of robots and changing action space.

In the ongoing work, we will be examining a larger scale deployment of robots using the methods examined here, and the impact of physical disturbances on the system. In the near future, we plan to theoretically analyze the coordination behavior of the sampling robots. One interesting idea would be to learn the number of robots needed to efficiently sample a given spatial field using prior knowledge about the field itself.





\bibliographystyle{spmpsci}      

\bibliography{bibliography.bib}

\end{document}